# ОБЗОР МЕТОДОВ НЕЙРОУПРАВЛЕНИЯ

Рассматриваются методы применения нейронных сетей для решения задач управления динамическими объектами. Для каждого вида нейроуправления приводятся схемы соединения нейросетей внутри системы управления и детально описываются процедуры их обучения. Анализируются преимущества и недостатки описанных методов.

## Введение

Нейроуправление динамическими объектами является новым перспективным направлением, находящимся на стыке таких дисциплин, как автоматическое управление, искусственный интеллект, нейрофизиология. Нейронные сети обладают рядом уникальных свойств, которые делают их мощным инструментом для создания систем управления: способностью к обучению на примерах и обобщению данных, способностью адаптироваться к изменению свойств объекта управления и внешней среды, пригодностью для синтеза нелинейных регуляторов, высокой устойчивость к повреждениям своих элементов в силу изначально заложенного в нейросетевую архитектуру параллелизма. В литературе описаны многочисленные примеры практического применения нейронных сетей для  решения задач управление самолетом [1–3], автомобилем [4], горнообогатительным процессом  [5], скоростью вращения вала двигателя [6],  электропечью [7], турбогенератором [8], сварочным аппаратом [9], пневмоцилиндром [10].

По-видимому, идея применения нейронных сетей для управления динамическими объектами впервые была высказана У. Видроу [11] еще в 1964 г., однако интенсивные исследования этого направления развернулись лишь в конце 80-х годов прошлого века. Один из первых обзоров в области нейроуправления  (1992 г.) упоминает 5 методов обучения нейросети для непосредственного управления объектом. С тех пор количество методов нейроуправление многократно увеличилось, появились новые решения на основе многомодульного подхода и приближенного динамического программирования.

В ходе развития нейроуправления, исследовались различные способы построения нейроконтроллеров с применением различных типов нейронных сетей: линейных типа «Адалина» [12], многослойных персептронов [13], рекуррентных сетей [14], сетей радиальных базисных функций [1] и др. Наилучшие результаты получены при использовании многослойных персептронов с линиями задержек [8], [15], [16]. Сформировались два основных направления в применения нейронных сетей внутри синтезируемых регуляторов: *прямые методы*, основанные на непосредственном управлении объектом с помощью нейронной сети,  и *непрямые методы*, когда нейронная сеть используется для выполнения вспомогательных функций управления, таких как фильтрация шума  или идентификация динамического объекта. В зависимости от числа нейронных сетей, составляющих нейроконтроллер, системы нейроуправления могут быть *одномодульными* или *многомодульными*. Схемы нейроуправления, которые применяются совместно с традиционными контроллерами, называются *гибридными*.

Ключевой проблемой при решении задач управления динамическими объектами является реализация модели инверсной динамики управляемого объекта. Аналитическое решение этой задачи не всегда возможно, поскольку требуется обращение причинно-следственных зависимостей по-ведения реального объекта. Применение нейронных сетей позволяет находить приближенные решения этой задачи путем обучения сети на примерах управления реальным объектом. При использовании прямых методов нейроуправленяяия, в частности, в методе обобщенного инверсного нейроуправления [7], [15], [17–20] это достигается путем непосредственного обучения нейронной сети на примерах поведения управляемого объекта. Однако, используемые для такого обучения последовательности примеров, полученные путем обращения результатов наблюдения реальных объектов часто содержат противоречия, резко снижающие качество обучения нейронной сети. Для решения этой проблемы предложен ряд методов. В методе специализированного инверсного нейроуправления [15], [17–21], [22] и некоторых версиях систем адаптивной критики [3] проблема обучения инверсной динамике решается путем аппроксимации аналитической модели управляемого объекта и вычисления локальных значений якобиана для различных областей пространства состояний. В методе обратного распространения ошибки через прямой нейроэмулятор для формирования линеаризованной модели инверсной динамики объекта используется обычная схема обратного распространения ошибки, применяемая для обучения многослойных персептронов. В системах многомодульного нейроуправлении эта же задача решается путем разделения пространства состояний объекта на локальные области, в которых инверсные модели представлены однозначными функциями. Для каждой такой области выделяется отдельный нейронный модуль [20], [23]. Перспективными для моделирования инверсной динамики могут оказаться новые типы нейронных сетей, позволяющие моделировать многозначные функции, в частности, вероятностные сети Бишопа на основе смесей гауссовских моделей (Mixture Density Networks) [24].

## 1. Структура системы управления динамическими объектами

В задачах нейроуправления для представления объекта управления используют модель черного ящика, в котором наблюдаемыми являются текущие значения входа и выхода. Состояние объекта считается недоступным для внешнего наблюдения, хотя размерность вектора состояний обычно считается фиксированной. Динамику поведения объекта управления можно представить в дискретном виде:

$$S(k+1) = \Phi(S(k), u(k)), \qquad (1)$$

$$y(k+1) = \Psi(S(k)), \qquad (2)$$

где $S(k) \in \Re^N$ – значение $N$-мерного вектора состояния объекта на $k$-м такте; $u(k) \in \Re^P$ – значение $P$-мерного вектора управления; $y(k+1) \in \Re^V$ – значение $V$-мерного выхода объекта управления на такте $k+1$.

Общая схема управления динамическим объектом показана на рис.1.

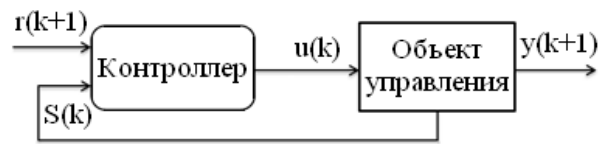

Рис. 1. Общая схема управления по обратной связи

Для оценки вектора состояния динамического объекта порядка может быть использована модель нелинейной авторегрессии с дополнительными входными сигналами (NARX) [25]:

$$S(k) = \begin{pmatrix} y(k) \\ y(k-1) \\ ... \\ y(k-N) \\ u(k-1) \\ u(k-2) \\ ... \\ u(k-Q) \end{pmatrix}. \qquad (3)$$

На практике, это соотношение обычно используют без ретроспективных управляющих входов:

$$S(k) = \begin{pmatrix} y(k) \\ y(k-1) \\ ... \\ y(k-N) \end{pmatrix}. \quad (4)$$

Состояние динамического объекта можно также представить мгновенным снимком его фазовой траектории:

$$S(k) = \begin{pmatrix} y(k) \\ y(k)' \\ ... \\ y(k)^{(N)} \end{pmatrix}. \quad (5)$$

При описании конкретных схем нейроуправления мы будем отдавать предпочтение модели (4). На схемах для ввода в контроллер задержанных данных обратной связи будет использоваться модуль линии задержек «TDL» (Tapped Delay Line). Для упрощения мы будем рассматривать только одноканальные системы управления (SISO), однако приводящиеся соотношения могут использоваться и для многоканальных систем (MIMO). Для этого следует лишь заменить в формулах скалярные переменные на входе и выходе объекта управления векторными.

## 2. Подражающее нейроуправление

Название «подражающее нейроуправление» (Neurocontrol learning based on mimic, Controller Modeling, Supervised Learning Using an Existing Controller) [11], [19–21] охватывает системы нейроуправления, в которых нейроконтроллер обучается на примерах динамики обычного контроллера по обратной связи, построенного, например, на основе обычной пропорционально-интегрально-дифференциальной (ПИД) схемы управления. Схема подражающего нейроуправления показана на рис. 2. После обучения нейронная сеть в точности воспроизводит функции исходного контроллера. В качестве примеров динамики контроллера может быть использована запись поведения человека-оператора. Обучающая выборка для нейронной сети формируется следующим образом.

Обычный контроллер по обратной связи (или человек-оператор) управляет объектом управления в штатном режиме. Значения величин на входе и выходе контроллера протоколируются, и на основе протокола формируется обучающая выборка для нейроной нейроной сети $U = \{P_i, T_i\}_{i=1}^{M}$, содержащая $M$ пар значений входа $P_i$ и ожидаемых реакций $T_i$ нейросети:

$$P_i = [r(i+1) \quad S(i)]^T, \quad (6)$$
$$T_i = u(i). \quad (7)$$

После обучения с помощью, например, метода обратного распространения ошибки, нейронная сеть подключается вместо исходного контроллера. Полученный нейроконтроллер может заменить человека в управлении устройством, а также быть более выгодным экономически, чем исходный контроллер. Основным недостатком этого метода является необходимость в предварительно настроенном исходном контроллере, что не всегда представляется возможным. Кроме того, полученный путем обучения нейроконтроллер в принципе не может обеспечить лучшее

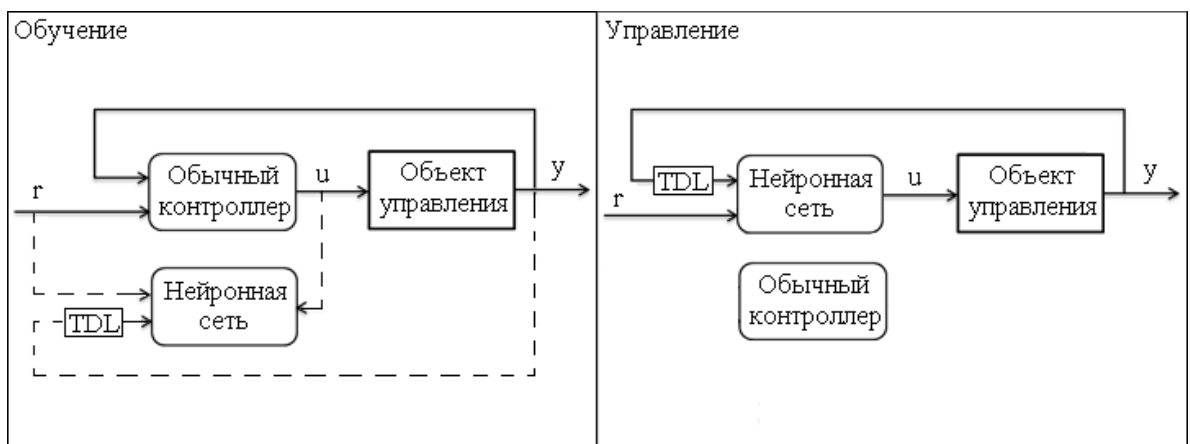

Рис. 2. Схема подражающего нейроуправления: слева – режим обучения нейронной сети; справа – режим управления качество управления, чем копируемый контроллер. Поэтому, сейчас подражающее нейроуправление применяют, в основном, для первичного обучения нейронной сети с использованием других методов для последующего дообучения нейроконтроллера.

### 3. Инверсное нейроуправление

При инверсном нейроуправлении формирование инверсной модели объекта управления осуществляется путем обучения нейронной сети. Известно несколько разновидностей такого нейроуправления.

**Обобщенное инверсное нейроуправление** (Generalized Inverse Neurocontrol, Direct Inverse Neurocontrol) [7], [15], [17–20], предусматривает обучение сети в режиме офф-лайн, на основе записанных траекторий поведения динамического объекта. Для получения таких траекторий, на объект управления в качестве управляющего сигнала подают некоторый случайный процесс. Значения управляющих сигналов и ответных реакций объекта протоколируют и на этой основе формируют обучающую выборку $U = \{P_i, T_i\}_{i=1}^{M}$:

$$P_i = [y(i) \quad S(i-1)]^T, \qquad (8)$$
$$T_i = u(i). \qquad (9)$$

В ходе обучения, нейронная сеть должна уловить и запомнить зависимость значений управляющего сигнала $u(k-1)$ от последующего значения реакции объекта управления $y(k)$, находящегося перед этим в состоянии $S(k-1)$. Для обучения нейронной сети используют метод обратного распространения ошибки (см. [24, 25]). Эту нейронную сеть называют «инверсный нейроэмулятор».

При управлении объектом, инверсный нейроэмулятор подключается как контроллер, причем возможны два способа подключения: замкнутый и разомкнутый.

При замкнутом подключении, схема которого показана на рис. 3 слева, на вход нейроконтроллера подаются текущие значения уставки и вектора состояния объекта управления, поступающего по цепи обратной связи:

$$x(k) = [r(k+1) \quad S(k)]^T. \qquad (10)$$

Благодаря стабилизирующему влиянию обратной связи, достигается достаточно высокое качество управления динамическим объектом. В работе [47] представлена вариация обобщенного инверсного управления, в которой в качестве уставки вместо одного целевого значения подается целевая траектория на $L$ тактов вперед: $[r(k+1) \quad r(k+2) \quad ... \quad r(k+L)]$.

При разомкнутом подключении на вход нейроконтроллера поступают только значения уставки с задержками:

$$x(k) = [r(k+1) \quad r(k) \quad .. \quad r(k-N+1)]^T \qquad (11)$$

При этом предполагается, что сформированная при обучении инверсная модель объекта управления является адекватной, следовательно сигнал управления, выдаваемый нейронной сетью, обеспечит переход объекта в положение, заданное уставкой. Разомкнутая система нейроуправления обладает высоким быстродействием, поскольку на вход нейроконтроллера не поступает значение текущего со-

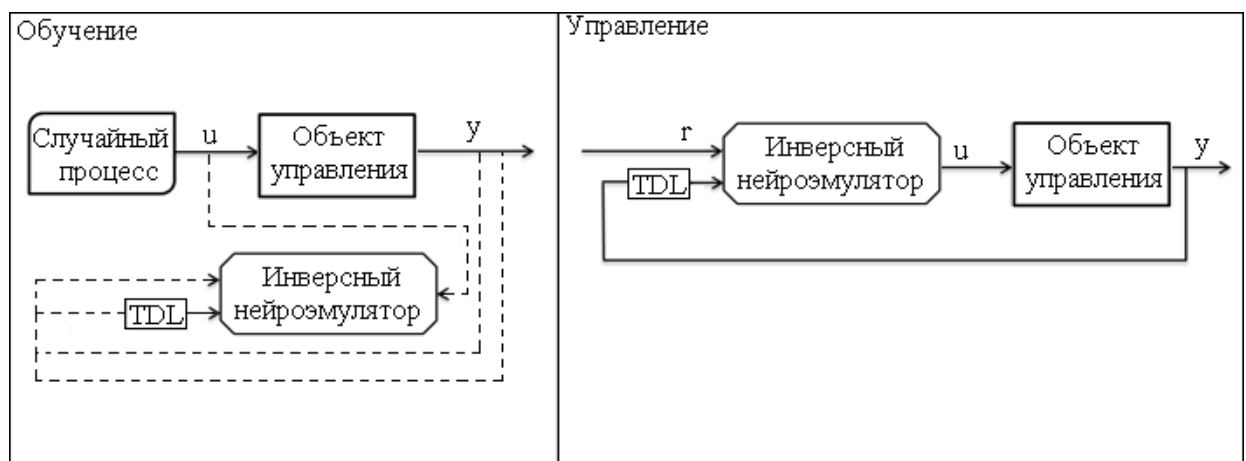

Рис. 3. Схема обобщенного инверсного нейроуправления: слева – режим обучения инверсного нейроэмулятора; справа – режим управления объектом стояния объекта управления, обработка которого требует значительных ресурсов. Однако, из-за отсутствия обратной связи качество такого управления оказывается низким [20].

Достоинством обобщенного инверсного нейроуправления является обучения нейроконтроллера в режиме офф-лайн и отсутствие необходимости в точной математической модели объекта управления. К недостаткам следует отнести сложность формирования обучающей выборки из-за необходимости тщательного подбора идентифицирующего случайного процесса, подаваемого на вход системы, а также низкое качество работы в тех случаях, когда инверсия объекта управления оказывается неоднозначной функцией. Неоднозначность приводит к наличию противоречий в обучающей выборке, заводящих в тупик процесс обучения нейронной сети.

**Специализированное инверсное нейроуправление** (Specialised Inverse Neurocontrol) [15], [17–22], позволяет обучать инверсный нейроконтроллер в режиме он-лайн, используя ошибку отклонения положения объекта от уставки $e = r - y$. Схема подключения нейронной сети к объекту управления показана на рис. 3, справа. На вход сети поступает вектор

$$x(k) = [r(k+1) \quad S(k)]^T. \qquad (12)$$

В ответ нейронная сеть генерирует управляющий сигнал $u(k)$, который приводит объект управления в положение $y(k+1)$. Далее вычисляется ошибка работы нейроконтроллера

$$e(k) = r(k+1) - y(k+1). \qquad (13)$$

Коррекция весовых коэффициентов нейронной сети выполняется по методу наискорейшего спуска:

$$w(k+1) = w(k) - \Delta w(k), \qquad (14)$$

$$\Delta w(k) = -\alpha \, e(k) \frac{\partial y(k+1)}{\partial u(k)} \frac{\partial u(k)}{\partial w(k)}. \qquad (15)$$

Здесь $\alpha$ — параметр скорости обучения нейронной сети. Величина производной в правой части формулы (14) $\frac{\partial u(k)}{\partial w(k)}$ вычисляется по методу обратного распространения ошибки. Производная $\frac{\partial y(k+1)}{\partial u(k)}$ представляет собой якобиан объекта управления, значение которого можно найти аналитически по заданной математической модели объекта управления. Однако, на практике, для получения приемлемого качества управления часто бывает достаточно вычислить лишь знак якобиана [22], [25]. Итерации коррекции значений коэффициентов $w$ продолжаются до достижения приемлемого качества управления.

Плюсом данного подхода является более высокое качество управления по сравнению с обобщенным методом инверсного нейроуправления. Его существенным недостатком является необходимость знания точной математической модели объекта управления, требуемой для обучения нейроконтроллера.

**Метод обратного пропуска ошибки через прямой нейроэмулятор** (Backpropagation Through Time, Internal Model Control) [17], [19], [20], [26–28] основан на идее применения тандема из двух нейронных сетей, одна из которых выполняет функцию контроллера, а вторая прямого нейроэмулятора, который обучается моделировать динамику объекта управления (рис. 4). В процессе обучения нейроконтроллера, текущая ошибка управления пропускается через нейроэмулятор в обратном направлении.

При обучении прямого нейроэмулятора, на вход объекта управления подается случайный управляющий сигнал $u$, изменяющий положение объекта управления $y$, и формируется обучающая выборка $U = \{P_i, T_i\}_{i=1}^{M}$:

$$P_i = [u(i) \quad S(i-1)]^T, \qquad (16)$$
$$T_i = y(i). \qquad (17)$$

Обучение прямого нейроэмулятора выполняется в режиме офф-лайн. Нейроэмулятор считается обученным, если при одинаковых значениях на входах нейро

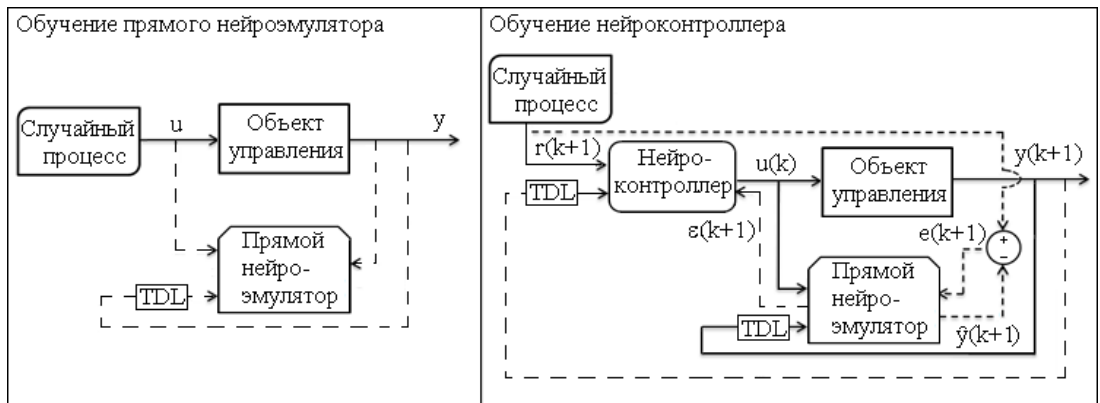

Рис. 4. Метод обратного пропуска ошибки через прямой нейроэмулятор: слева – схема обучения прямого нейроэмулятора; справа – схема обучения нейроконтроллера эмулятора и реального объекта, отличие между значениями их выходов становится незначительным. После завершения обучения прямого нейроэмулятора, проводится обучение нейроконтроллера. Обучение выполняется в режиме он-лайн по такой же схеме, как и в случае инверсного специализированного нейроуправления. Сначала (на такте $k$) на вход нейроконтроллера поступает желаемое положение объекта управления для следующего такта $r(k+1)$. Нейроконтроллер генерирует сигнал управления $u(k)$, который поступает на входы объекта управления и нейроэмулятора. В результате, управляемый объект переходит в положение $y(k+1)$, а нейроэмулятор генерирует реакцию $\hat{y}(k+1)$. Далее вычисляется ошибка управления $e(k) = \hat{y}(k+1) - y(k+1)$ и пропускается в обратном направлении по правилу обратного распространения. Весовые коэффициенты связей нейроэмулятора при этом не корректируются. Механизм обратного прохождения ошибки через прямой нейроэмулятор реализует локальную инверсную модель в текущей точке пространства состояний объекта управления. Пройдя через нейроэмулятор, ошибка далее распространяется через нейроконтроллер, но теперь ее прохождение сопровождается коррекцией весовых коэффициентов нейроконтроллера. Нейроэмулятор при этом выполняет функции дополнительных слоев нейроной сети нейроконтроллера, в которых веса связей не корректируются.

## 4. Прогнозирующее нейроуправление

Метод обучения нейроконтроллеров, при котором минимизируется отклонение текущего положения объекта управления от уставки для каждого такта, не всегда обеспечивает наилучшее интегральное качество управления, оцениваемое выражением:

$$IAE = \sum_{k=1}^{K}(r(k) - y(k))^2. \qquad (18)$$

Причин тому несколько. Во-первых, качество управления ухудшается из-за свойства запаздывания минимум на один такт, общего для систем управления по обратной связи. Во-вторых, если для достижения целевого положения нужно несколько тактов, нейроконтроллер, стремясь минимизировать текущую ошибку, может выдать чрезмерно сильный управляющий сигнал, что ведет к перерегулированию.

**Прогнозирующее модельное нейроуправление** (NN Predictive Control, Model Predictive Control, Neural Generalized Predictive Control) [17], [29], [30–33] минимизирует функционал стоимости интегральной ошибки, прогнозируемой на $L = \max(L_2, L_3)$, $0 \le L_1 \le L_2$ тактов вперед:

$$Q(k) = \sum_{i=L_1}^{L_2} e(k+i)^2 + \\ + \rho \sum_{i=0}^{L_2}(u(k+i) - u(k+i-1))^2. \qquad (19)$$

Здесь $e$ – ошибка выхода системы, $\rho$ – вклад изменения управляющего сигнала в общий функционал стоимости $Q$. Схема показана на рис. 5. Для прогнозирования будущего поведения системы и вычисления ошибок используется прямой нейроэмулятор, обученный так же, как в случае обратного распространения ошибки через прямой нейроэмулятор (см. рис. 4, слева). Примечательность этого метода состоит в том, что в нем отсутствует обучаемый нейроконтроллер. Его место занимает оптимизационный модуль, работающий в режиме реального времени, в котором может быть использован сиплекс-метод [31] или квази-Ньютоновский алгоритм [32].

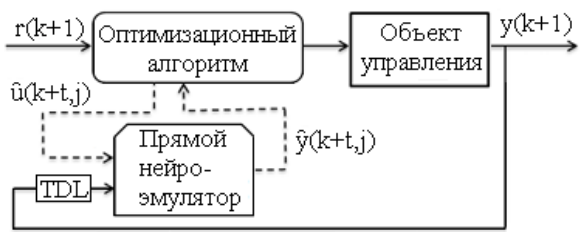

Рис. 5. Схема прогнозирующего модельного нейроуправления

Оптимизационный модуль получает на такте $k$ целевую траекторию на $L$ тактов вперед, а если ее нет, то $L$ раз дублирует значение текущей уставки $r(k+1)$ и использует это в качестве целевой траектории. Далее, для выбора оптимального управляющего воздействия, вычисления происходят во внутреннем цикле системы нейроуправления (его итерации обозначены как $j$). За время одного такта управления оптимизационный модуль подает на вход нейроэмулятора серию различных воздействий $\hat{u}(k+t, j)$, где $t$ – глубина прогнозирования, $0 \le t \le L-1$, получает различные варианты поведения системы $\hat{y}(k+t+1, j)$, вычисляет функцию стоимости по формуле (19) и определяет наилучшую стратегию управления $ST = \{\hat{u}(k, j_1)\ \ \hat{u}(k+1, j_2)\ ;...;\ \hat{u}(k+L-1, j_L)\}$ в смысле минимизации функционала стоимости (19). В итоге, на объект подается управляющий сигнал $u(k) = \hat{u}(k, j_1)$. На следующем такте стратегия $ST$ пересчитывается заново.

Минусом систем прогнозирующего модельного нейроуправления является невозможность их применения в системах с большой частотой дискретизации, так как оптимизационный алгоритм, работающий в режиме реального времени, за время одного такта не будет успевать находить наилучшую стратегию действий.

**Методы нейроуправления на основе адаптивной критики** (Adaptive Critics), которые также известны как «Приближенное динамическое программирование» (Approximated Dynamic Programming, ADP), в последние годы весьма популярны [3], [8], [16], [33–35]. Подобно системам прогнозирующего модельного управления, системы адаптивной критики выбирают управляющий сигнал на основе оценок ошибок будущего с бесконечным горизонтом:

$$J(k) = \sum_{i=0}^{\infty} \gamma^i e(k+i)^2. \qquad (20)$$

Здесь $\gamma$ – коэффициент забывания, $0 < \gamma \le 1$; $e(k)$ – ошибка, вычисляемая по формуле (13). Система включает два нейронных модуля: нейроконтроллер и модуль критики. Нейроконтроллер обучают минимизировать функционал стоимости $J(k)$, который играет ту же роль, что и ошибка $e(k)$ в методах обучения по ошибке обратной связи. Модуль критики выполняет аппроксимацию значений функции стоимости.

На рис. 6, слева показана схема работы системы адаптивной критики в режиме управления объектом. На вход нейроконтроллера поступает вектор $x(k) = [r(k+1)\ \ S(k)]^T$, вызывающий появление на его выходе сигнала управления $u(k)$, в результате чего объект управления переходит в положение $y(k+1)$. Далее производится вычисление значения текущей ошибки управления $e(k)$.

Модуль критики, получая на входе вектор $z(k) = [r(k+1)\ \ u(k)\ \ S(k)]^T$, производит оценку функции стоимости $J(k)$. На следующем такте процесс повторяется:

вычисляются новые значения $e(k+1)$ и $J(k+1)$.

Обучение системы нейроуправления происходит в режиме он-лайн и состоит из двух этапов: обучения модуля критики и обучения нейроконтроллера. Сначала, рассчитывается ошибка временной разности $\delta(k)$:

$$\delta(k) = e(k) + \gamma \, \hat{J}(k+1) - \hat{J}(k). \quad (21)$$

Затем по методу наискорейшего спуска выполняется коррекция веса связей для модуля критики $w_{CRITIC}$:

$$\Delta w_{CRITIC}(k) = -\alpha_1 \delta(k) \frac{\partial J(k)}{\partial w_{CRITIC}(k)}. \quad (22)$$

Значение градиента $\frac{\partial J(k)}{\partial w_{CRITIC}(k)}$ рассчитывается по методу обратного распространения ошибки. Коррекция веса связей нейроконтроллера $w_{CONTROL}$ производится аналогично:

$$\Delta w_{CONTROL}(k) = -\alpha_2 \frac{\partial J(k)}{\partial u(k)} \frac{\partial u(k)}{\partial w_{CONTROL}(k)}. \quad (23)$$

Значение производной находят путем обратного распространения величины через модуль критики, а значение градиента – путем обратного распространения ошибки через модуль контроллера. Коррекция весов продолжается, пока система не достигнет требуемого уровня качества управления. Таким образом, на каждом шаге улучшается закон управления, путем обучения нейроконтроллера (Policy Iteration), а также повышается способность системы оценивать ситуацию, путем обучения критика (Value Iteration).

Конкретная схема построения системы адаптивной критики может отличаться от вышеописанной, носящей название Heuristic Dynamic Programming (HDP). В методе DHP (Dual Heuristic Programming), где модуль критики вычисляет производную функционала глобальной стоимости $\frac{\partial J}{\partial t}$, а в методе GDHP (Global Dual Heuristic Programming) вычисляются как сам функционал функции стоимости $J$, так и его производная $\frac{\partial J}{\partial t}$.

Известны модификации метода, в которых модуль критики принимает решения исключительно на основе управляющего сигнала $u(k)$ [35]. Они имеют приставку AD («Action Dependent»): ADHDP, ADDHP, ADGDHP. В некоторых версиях адаптивной критики модуль критики состоит из двух частей: собственно, модуля критики и прямого нейроэмулятора. Последний выдает предсказания поведение объекта управления, на основе которых критик формирует оценку функции стоимости $J$. Такие версии носят название «основанные на модели» («Model based»). Систематическое описание существующих разновидностей систем адаптивной критики представлено в [3].

Популярность систем адаптивной критики объясняется наличием развитой теоретической базы в виде теории динамического программирования Беллмана, а также их способностью сходиться к оптимальному или близкому к оптимальному управлению [35].

## 5. Многомодульное нейроуправление

Многомодульные нейросистемы, построенные по типу комитетов экспертов [25], получили значительное распространение в системах распознавания, позже они дали толчок развитию многомодульных систем нейроуправления. В рамках многомодульного подхода, исходная задача разделяется на отдельные подзадачи,

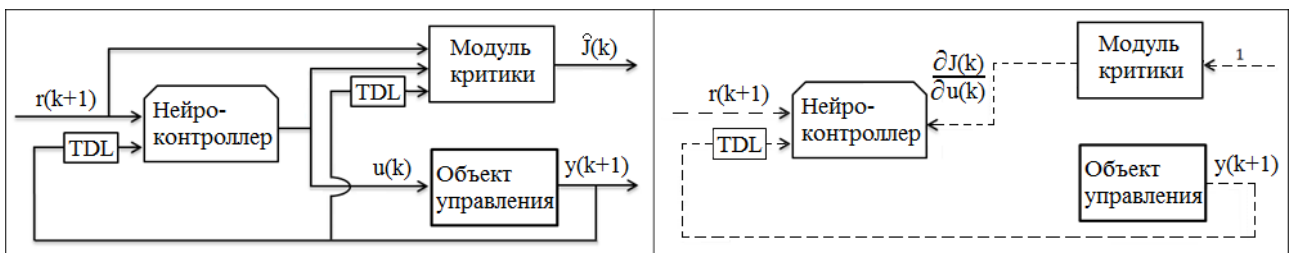

Рис. 6. Схема адаптивной критики: слева – этап управления; справа – этап обучения которые решают отдельные модули. Финальное решение выполняет шлюзовая сеть на основе частных решений модулей-экспертов.

**Системы многомодульного нейроуправления на основе локальных инверсных моделей** (Incremental Clustered Control Networks) [20], [23], состоят из множества линейных нейроконтроллеров и шлюзового модуля. Каждый из линейных нейроконтроллеров представляет нейронную сеть Адалина [25], обученную управлять в пределах локальной области пространства состояний объекта:

$$LS_l = \begin{pmatrix} (y_{l1} - d_{l1}; y_{l1} + d_{l1}) \\ ... \\ (y_{lN} - d_{lL}; y_{lL} + d_{lL}) \end{pmatrix}.$$

Здесь $L$ – количество модулей-экспертов. Преимуществом линейных сетей перед многослойными персептронами состоит в том, что их поведение проще анализировать, а также они быстрее обучаются. Это особенно важно для анализа устойчивости синтезируемой системы управления. Для формирования линейных нейроконтроллеров могут применяться различные методы: обобщенное инверсное нейроуправление, специализированное инверсное нейроуправление, метод обратного пропуска ошибки через нейроэмулятор.

После того, как локальные линейные нейроконтроллеры были обучены, производится обучение шлюзовой сети. Она обучается по входной оценке состояния управляемого объекта $S(k)$ находить локальную область пространства состояний $LS_j$ такую, что $S(k) \in LS_j$ и выдавать на объект управления сигнал $u_j(k)$, сгенерированный локальным линейным нейроконтроллером, соответствующим этому локальному участку. Недостатком этого метода является необходимость в большом количестве примеров для обучения нейроконтроллеров, распределенных во всех областях пространства состояний управляемого объекта.

**Метод многомодульного нейроуправления на основе пар прямых и инверсных моделей** (Multiple Paired Forward and Inverse Models, Multiple Switched Models), [36–40] показан на рис. 7. В отличие от метода нейроуправления на основе локальных инверсных моделей, в котором поведение системы формируется при обучении, и в ходе управления не корректируется, данный метод предусматривает корректировку поведения нейронных модулей на каждом такте нейроуправления. Для этого, каждый модуль включает два нейроэмулятора: прямой и инверсный. Обучение прямого нейроэмулятора производится по схеме метода обратного пропуска ошибки через прямой нейроэмулятор, показанной на рис. 4, слева. Инверсный нейроэмулятора обучается по схеме обобщенного инверсного нейроуправления, показанной рис. 3, слева. Предполагается, что каждая пара нейроэмуляторов обучается на своем примере динамики объекта управления

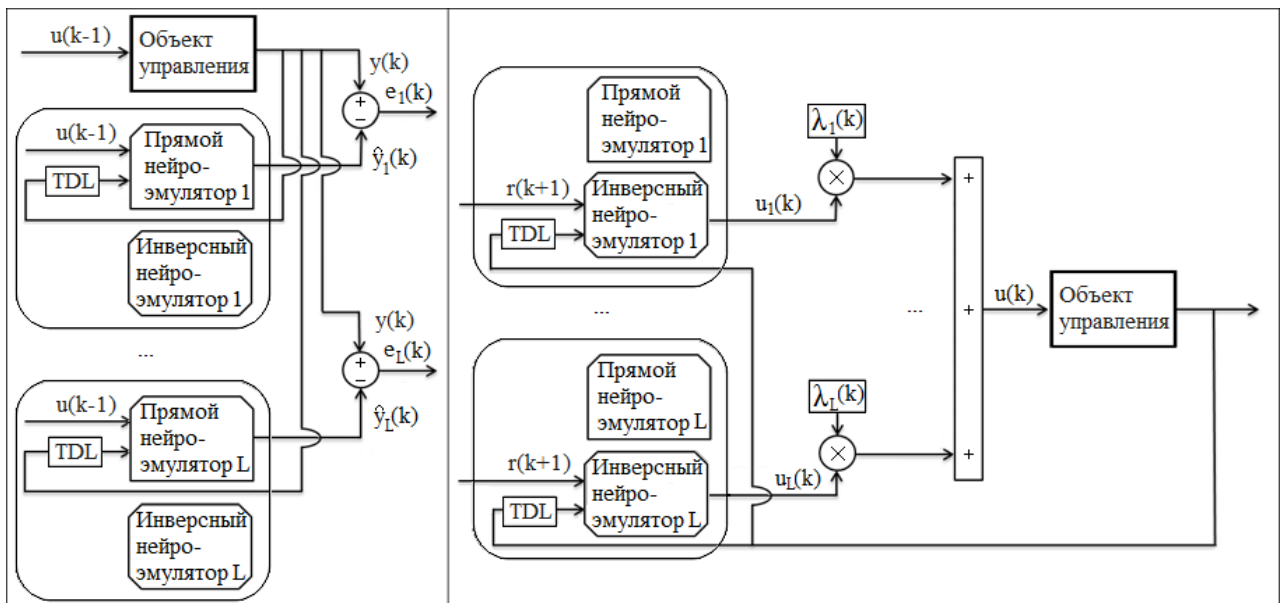

Рис. 7. Схема многомодульного нейроуправления на основе пар прямых и инверсных моделей: слева – этап переоценки коэффициентов ответственности модулей; справа – этап коллективного управления и специализируется именно на нем. Поэтому, если прямой нейроэмулятор правильно предсказывает динамику объекта управления, то соответствующий ему инверсный нейроэмулятор хорошо управляет объектом. Предполагается также, что применяющиеся для обучения пар эмуляторов траектории состояний управляемого объекта существенно отличаются между собой.

Работа системы на каждом такте включает два этапа: 1) переоценки коэффициентов ответственности модулей и 2) коллективного управления модулями на основе вычисленных коэффициентов ответственности. Общая схема работы системы нейроуправления, состоящей из $L$ модулей, показана на рис. 7. На первом этапе, на вход прямого нейроэмулятора каждого из модулей поступает сигнал $u(k-1)$, соответствующий значению управления на предыдущем такте, а также вектор предыдущего состояния $S(k-1)$, характеризующий предыдущее положение управляемого объекта. По входным данным, каждый прямой нейроэмулятор производит свою оценку текущего положения объекта $\{\hat{y}_1(k) \ ; \ ... \ ; \ \hat{y}_L(k)\}$, после чего вычисляются ошибки оценок предвидения для всех модулей системы:

$$\{e_1(k) \ ; \ ... \ ; \ e_L(k)\},$$
$$e_l(k) = y(k) - \hat{y}_l(k), 1 \le l \le L. \quad (24)$$

На основе ошибок предвидения, рассчитываются коэффициенты предвидения $\{\lambda_1(k) \ ; \ ... \ ; \ \lambda_L(k)\}$, $\sigma$ – масштабирующая константа:

$$\lambda_l(k) = \frac{\exp(-e_l(k)^2/\sigma^2)}{\sum_{j=1}^{L}\exp(-e_j(k)^2/\sigma^2)},$$
$$\sum_{j=1}^{L}\lambda_j = 1. \quad (25)$$

На этапе управления, инверсный нейроэмулятор $l$-го модуля действует по схеме обобщенного инверсного нейроуправления. На его вход поступают значение уставки $r(k)$ и оценки текущего состояния объекта $S(k-1)$, вызывая реакцию $u_l(k)$. Итоговый управляющий сигнал $u(k)$ представляет собой взвешенную сумму управляющих сигналов отдельных модулей, при этом управляющий сигнал каждого модуля обеспечивает вклад, пропорциональный коэффициенту предвидения соответствующего модуля:

$$u(k) = \sum_{l=1}^{L}\lambda_l(k)u_l(k). \quad (26)$$

В некоторых системах, вместо этой формулы при выборе текущего управляющего модуля применяют принцип «победитель получает все» [35, 36]. Впрочем, проблема выбора способа декомпозиции задачи на подзадачи характерна для многомодульного подхода вообще.

Существенным минусом систем многомодульного нейроуправления является непрозрачная процедура разделения обучающей выборки на подвыборки для обучения прямых и инверсных нейроэмуляторов разных модулей.

## 6. Гибридное нейроуправление

Гибридными называют системы нейроуправления, в которых нейронные сети работают совместно с обычными контроллерами, ПИД-регуляторами или другими типами контроллеров.

**Гибридное нейро-ПИД управление** (NNPID Auto-tuning, Neuromorphic PID Self-tuning) [9], [17], [41, 43] позволяет осуществлять самонастройку ПИД-регулятора в режиме он-лайн с использованием нейронных сетей.

Управление с использованием ПИД-контроллера основано на минимизации ошибки обратной связи. Вырабатываемый контроллером сигнал управления представляет взвешенную сумму пропорциональной, интегральной и дифференциальной частей:

$$u(t) = K_1 e(t) + K_2 \int_0^t e(\tau)d\tau + K_3 \frac{de(t)}{dt}. \quad (27)$$

Коэффициенты $K_1$, $K_2$, $K_3$ получаются при настройке ПИД-контроллера, которая может быть выполнена вручную по правилу Зиглера – Никольса, правилу Коэна – Куна или другими методами [43], либо с использованием нейронной сети, как показано на рис. 8.

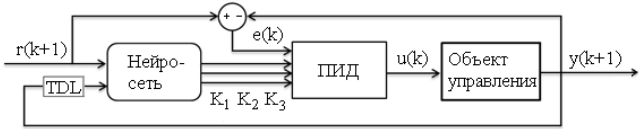

Рис. 8. Схема гибридного нейро-ПИД управления

Обученная система нейроуправления действует следующим образом. На такте $k$ нейронная сеть получает уставку $r(k+1)$ и генерирует коэффициенты управления ПИД-контроллера $K_1(k)$, $K_2(k)$, $K_3(k)$, которые поступают на ПИД-контроллер вместе со значением текущей ошибки обратной связи $e(k)$, вычисляемой по формуле (11). ПИД-контроллер рассчитывает управляющий сигнал $u(k)$ по формуле:

$$u(k) = u(k-1) + K_1(k)(e(k)-e(k-1)) + \\ + K_2(k)e(k) + K_3(k)(e(k) - \\ - 2e(k-1) + e(k-2)), \quad (28)$$

применяемой для дискретных ПИД-контроллеров и подает его на объект управления.

Обучение нейросети происходит в режиме реального времени по ошибке обратной связи, методом наискорейшего спуска:

$$\Delta w(k) = -\alpha\, e(k) \frac{\partial y(k+1)}{\partial u(k)} \frac{\partial u(k)}{\partial K(k)} \frac{\partial K(k)}{\partial w(k)}. \quad (29)$$

Здесь $K(k) = [K_1(k)\ K_2(k)\ K_3(k)]^T$ – вектор выходов нейронной сети, поступающий на ПИД-контроллер.

$$\frac{\partial u(k)}{\partial K_i} = \begin{cases} e(k) - e(k-1) & \text{при } i=1; \\ e(k) & \text{при } i=2; \\ e(k) - 2e(k) + e(k-2) & \text{при } i=3. \end{cases} \quad (30)$$

Градиенты вычисляют методом обратного распространения ошибки. Якобиан находится аналитически, на основе математической модели объекта управления. Плюсами использования этого подхода является упрощение эксплуатации вследствие устранения процедуры настройки ПИД-контроллера вручную. Кроме того, в случае применения нейронной сети с нелинейными активационными функциями, ПИД-контроллер фактически превращается в нелинейный контроллер, что потенциально обеспечивает более высокое качество управления нелинейными динамическими объектами. Обратная сторона медали – сложность оценки устойчивости полученного нелинейного контроллера. Также минусом является необходимость в точной математической модели объекта управления, необходимой для вычисления якобиана объекта управления. Эту трудность можно обойти, используя прямой нейроэмулятор и действуя по методу обратного распространения ошибки через прямой нейроэмулятор.

**Методы гибридного параллельного нейроуправления** (Parallel Neurocontrol, Stable Direct Adaptive Control, NARMA L2 Feedback Linearization Control, Additive Feedforward Control) [7], [17], [29] предусматривают параллельное использование нейроконтроллеров и обычных контроллеров для управления динамическими объектами. Соответствующая схема показана на рис. 9. При этом нейроконтроллер и обычный контроллер, в роли которого выступает, например, ПИД-контроллер, получают одинаковые значения уставки.

Возможны следующие варианты совместного подключения обычного контроллера и нейроконтроллера:

1) к объекту управления подключается обычный контроллер, после чего нейроконтроллер обучается управлять уже замкнутой обычным контроллером систе-

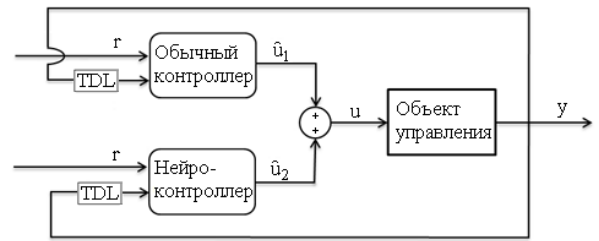

Рис. 9. Схема гибридного параллельного нейроуправления мой. После обучения нейроконтроллера, он подключается к системе, а управляющие сигналы обоих контроллеров суммируются;

2) нейроконтроллер учится управлять объектом управления, после обучения

начинает функционировать в штатном режиме. Далее, для управления замкнутой нейроконтроллером системой настраивается обычный контроллер. После настройки обычного контроллера, он подключается к системе, управляющий сигнал обоих контроллеров суммируется;

3) области действия обычного контроллера и нейроконтроллера разграничиваются. Например, в пространстве состояний объекта управления для нейроконтроллера выделяется отдельная область $LS$:

$$LS = \begin{pmatrix} (y_1 - d_1; y_1 + d_1) \\ ... \\ (y_N - d_N; y_N + d_N) \end{pmatrix}.$$

При этом, обычный контроллер рассчитывается на управление объектом вне этой области пространства состояния. При параллельной работе обоих контроллеров, управляющий сигнал поступает на объект либо от нейроконтроллера, если текущее состояние системы находится в пределах области $LS$, либо, в противном случае, от обычного контроллера.

Гибридное параллельное нейроуправление представляет компромиссное решение для внедрения нейроуправления в промышленность и перехода от обычных контроллеров к нейросетевым.

## 7. Вспомогательное нейроуправление

Нейронные сети могут решать различные вспомогательные задачи, возникающие в ходе управления динамическим объектом. Качество управления контроллера можно повысить и сделать траекторию движения объекта управления более гладкой при использовании **метода нейросетевой фильтрации внешних возмущений** (Adaptive Inverse Control, Adaptive Inverse Control based on Linear and Nonlinear Adaptive Filtering, Internal Model Control) [7], [12, 13], [18]. Изначально, эта схема была предложена Б. Видроу для использования совместно с нейроконтроллерами, обученными по методу обобщенно-инверсного нейроуправления [12]. В более поздней работе [13] им были применены нейроконтроллеры, обученные по методу обратного распространения ошибки через прямой нейроэмулятор. В принципе, нейросетевую фильтрацию ошибок можно использовать для повышения качества работы контроллера любого типа, не обязательно нейросетевого. Схема работы такой подсистемы показана на рис. 10. В ней используется две предварительно обученных нейронных сети: инверсный нейроэмулятор, обученный так же, как это делается в методе обобщенного инверсного нейроуправления (рис. 3, слева) и прямой нейроэмулятор, обученный так же, как это делается в методе обратного распространения ошибки через прямой нейроэмулятор (рис. 4, слева).

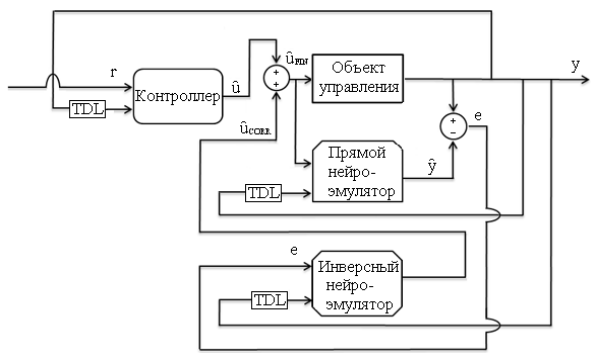

Рис. 10. Схема метода нейросетевой фильтрации внешних возмущений

Пусть на объект управления поступает управляющий сигнал $\hat{u}_{FIN}(k)$, явившийся результатом суммирования сигнала контроллера $\hat{u}(k)$ и корректирующего сигнала системы фильтрации внешних возмущений $\hat{u}_{CORR}(k)$, вычисленного на предыдущем такте. Сигнал $\hat{u}_{FIN}(k)$ направляется на прямой нейроэмулятор объекта управления, а реакция прямого нейроэмулятора $\hat{y}(k)$ сравнивается с реальным положением системы $y(k)$. Разница этих величин $e(k)$ трактуется как нежелательное отклонение системы, вызванное внешним возмущением. Для подавления нежелательного эффекта, сигнал $e(k)$ поступает на инверсный нейроэмулятор, который рассчитывает корректирующий сигнал $\hat{u}_{CORR}(k+1)$ для корректировки управляющего сигнала нейроконтроллера $\hat{u}(k+1)$ на следующем такте. Для использования этого метода, объект управления должен об-ладать обращаемой динамикой, а также необходимо иметь адекватную математическую или имитационную модель объекта управления для обучения прямого и инверсного нейроэмуляторов.

**Нейроуправление с эталонной моделью** (Model Reference Adaptive Control, Neural Adaptive Control) [11], [15], [18], [26] – вариант нейроуправления по методу обратного распространения ошибки через прямой нейроэмулятор, с дополнительно внедренной в схему эталонной моделью (Reference Model). Это делается в целях повышения устойчивости переходного процесса: в случае, когда переход объекта в целевое положение за один такт невозможен, траектория движения и время осуществления переходного процесса становятся плохо прогнозируемыми величинами и могут привести к нежелательным режимам работы системы. Схема нейроуправления с эталонной моделью показана на рис. 11.

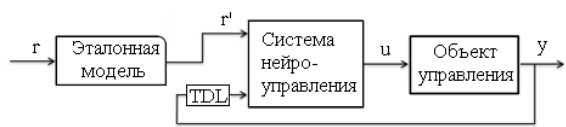

Рис. 11. Схема нейроуправления с эталонной моделью

Для уменьшения этой неопределенности, между уставкой и нейроконтроллером вводится эталонная модель, представляющая собой, как правило, линейную динамическую систему невысокого порядка, которую можно легко аналитически проверить на устойчивость. В ходе как обучения, так и управления, эталонная модель получает на вход уставку $r$ и генерирует опорную траекторию $r'$, которая дальше поступает на нейроконтроллер в качестве новой уставки, которую нужно выполнить. Эталонная модель подбирается таким образом, чтобы генерируемая ею опорная траектория на каждом такте была достижима для объекта управления.
Хотя под системой нейроуправления по эталонной модели чаще всего подразумевается именно система конструкции К. Нарендры и К. Пасарати [26], нет принципиальных ограничений против применения эталонных моделей совместно с системами нейроуправления других типов, например, обобщенного инверсного нейроуправления или специализированного инверсного нейроуправления.

К. Кришнакумаром и др. [2], [42] предложена оригинальная модификация нейроуправления с адаптируемой эталонной моделью для создания аварийно-устойчивой системы управления летательным аппаратом. В качестве контроллера используется классический неадаптируемый ПИД-контроллер, а эталонная модель представляет система нейроуправления типа адаптивной критики, способная менять свое поведение в ходе полета, генерируя на выходе для отслеживания контроллером различные опорные траектории. Эталонная модель дообучается в режиме он-лайн путем минимизации среднеквадратичной ошибки отклонения траектории движения объекта управления от целевой траектории. Такую адаптивную систему можно рассматривать как обычный нейроконтроллер типа адаптивной критики, управляющий объединенной динамической системой «ПИД-контроллер + объект управления».

**Выводы**

За последние 20 лет нейроуправление получило значительное развитие. Как было обозначено в одном из первых обзоров по тематике нейроуправления [21] в качестве перспективного направления исследований, доминирующая доля внимания была уделена задачам разработки нейросистем для управления нелинейными динамическими объектами, получено множество примеров успешно работающих систем этого типа. В качестве универсального эффективного метода нейроуправления был заявлен разработанный относительно недавно метод адаптивной критики. Показано, что рекуррентные сети типа NARX наилучшим образом подходят для моделирования динамических систем [44], что привело к их распространению в качестве идентификаторов объектов управления в непрямых и прогнозирующих методах нейроуправления. Вместе с тем, все существующие на сегодняшний день алгоритмы обучения рекуррентных нейросетей являются вариациями разработанных в начале 90-х алгоритмов BPTT и RTRL [45], обучение которых проходит сравнительно медленно и требует значительных вычислительных ресурсов.

Вместе с тем, остается ряд нерешенных проблем, мешающих широкому применению систем нейроуправления в индустрии.

1. Все еще отсутствует универсальная процедура анализа устойчивости нелинейных нейроконтролеров. Были предложены лишь частные решения для отдельных видов нейроконтроллеров при известной математической модели объекта управления.

2. Конструкция почти всех схем нейроуправления выглядит слишком усложненной из-за наличия нескольких нейросетей и нетривиальной последовательности процедур их обучения. Перспективным направлением исследований является получение унифицированного алгоритма обучения единой управляющей нейросети.

3. Для дальнейшего развития методов нейроуправления, актуальной проблемой остается создание новых моделей динамических нейронных сетей и способов из обучения, так как базовыми блоками построения многих методов нейроуправления являются эмпирически полученные модели прямой или инверсной динамики объекта управления.

*Об авторах:*

*Чернодуб Артем Николаевич,*
младший научный сотрудник,

*Дзюба Дмитрий Александрович,*
младший научный сотрудник.

*Место работы авторов:*

Институт проблем математических машин и систем НАН Украины,
03680, Киев-187,
Проспект Академика Глушкова, 40.
Тел.: (044) 526 5548.
E-mail: a.chernodub@gmail.com
          ddziuba@immsp.kiev.ua